\documentclass[10pt,twocolumn,letterpaper]{article}

\usepackage[final]{cvpr}
\usepackage{times}
\usepackage{graphicx}
\usepackage{booktabs}
\usepackage{amsmath,amssymb}
\usepackage[pagebackref,breaklinks,colorlinks]{hyperref}
\usepackage{xcolor}
\usepackage[font=small,labelfont=bf]{caption}
\usepackage{microtype}
\usepackage{multirow}
\usepackage{tabularx}
\usepackage{pifont}
\usepackage{enumitem}
\setlist{nosep,leftmargin=1.2em}
\def\paperID{****}
\def\confName{CVPR}
\def\confYear{2026}
\title{HalluCXR: Benchmarking and Mitigating Hallucinations in Medical Vision-Language Models for Chest Radiograph Interpretation}

\author{
Haoyu wang\textsuperscript{1} \quad Zitong Li\textsuperscript{1}\\
\textsuperscript{1}Department of Biostatistics \& Health Informatics,\\
Institute of Psychiatry, Psychology \& Neuroscience,\\
King's College London, London, UK\\
{\tt\small haoyu.7.wang@kcl.ac.uk, zitong.2.li@kcl.ac.uk}
}

\begin{document}
\maketitle

\begin{abstract}
Vision-language models (VLMs) are increasingly used for medical image interpretation, yet they frequently hallucinate, generating clinically plausible but factually incorrect findings that pose direct patient safety risks.
We introduce \textbf{HalluCXR}, a benchmark evaluating six architecturally diverse VLMs across 856 stratified MIMIC-CXR chest radiographs and three query types, yielding 15,408 model evaluations.
An eight-category hallucination taxonomy with clinical severity ratings and a two-layer detection pipeline are validated against 250 human annotations (auto-detection F1$=$0.959; LLM judge F1$=$0.907).
We find that 61.9--82.3\% of outputs contain hallucinations, with clinically dangerous errors in up to 80.2\%. Three key patterns emerge: normal radiographs paradoxically attract the most severe hallucinations, common findings are systematically over-fabricated while rare findings go under-detected, and response length alone predicts hallucination risk (AUC up to 0.908).
A six-model ensemble reduces fabrication by up to 84.8\% at the cost of increased omission; a three-model subset retains comparable performance at half the cost. These results establish that hallucination auditing, verbosity-based risk monitoring, and ensemble-based safety layers are prerequisites for clinical deployment.
\end{abstract}

\section{Introduction}

When asked to interpret a normal chest radiograph, five out of six leading vision-language models fabricate pathological findings, reporting near-perfect confidence while doing so.

This finding, drawn from 15,408 model outputs across six architecturally diverse VLMs, exposes a gap between the surface-level fluency of medical VLMs and their faithfulness to image content. Models such as GPT-5.1, Gemini 3 Flash~\cite{google2024gemini}, Qwen2-VL~\cite{wang2024qwen2vl}, InternVL2~\cite{chen2024internvl2}, and LLaVA-NeXT~\cite{liu2024llavanext} can produce detailed radiology reports~\cite{li2023llavamed,chen2024chexagent}, answer visual questions~\cite{moor2023medflamingo}, and support clinical decisions~\cite{yang2024gpt4v}. The outputs read well. But these models regularly \emph{hallucinate}: they fabricate pathological findings absent from the image or omit abnormalities visible to radiologists, with potentially life-threatening consequences. A fabricated pneumothorax on a normal radiograph could trigger an unnecessary chest tube insertion; an omitted pleural effusion could delay drainage.

Prior benchmarks either lack visual grounding~\cite{umapathi2023medhalt}, evaluate single models without cross-model comparison~\cite{gu2024medvh}, or target general-domain objects rather than radiological findings with clinical severity~\cite{rohrbach2018chair,li2023pope,jing2024faithscore}. Most treat hallucination as binary, without distinguishing fabrication from omission, and few propose mitigation beyond model retraining.

We address these gaps with the following contributions:
\begin{enumerate}
    \item \textbf{HalluCXR Benchmark}: A systematic evaluation of six architecturally diverse VLMs on 856 MIMIC-CXR radiographs across three query types, yielding 15,408 evaluations stratified by diagnostic complexity into four clinical strata.
    \item \textbf{Hallucination Taxonomy \& Detection}: An eight-category taxonomy (A1--C3) with a two-layer detection pipeline (negation-aware auto-detection plus LLM-as-judge cross-validation) validated against 250 human expert annotations (F1$=$0.959, $\kappa{=}$0.870).
    \item \textbf{Diagnostic Analysis}: Discovery that common findings are systematically over-fabricated while rare findings go under-detected, that normal radiographs attract the most severe errors, and that \textbf{response verbosity alone predicts hallucination risk} without requiring ground truth.
    \item \textbf{Mitigation Framework}: A multi-layer mitigation approach combining (a)~length-based risk triggers for real-time safety filtering, and (b)~ensemble strategies (calibration-weighted and label-aware voting) that reduce fabrication by up to 84.8\% and improve F1 by 8.7\% over simple voting, without retraining.
\end{enumerate}

\section{Related Work}

\subsection{Medical Vision-Language Models}

Medical VLMs fall into three broad paradigms. \textbf{CLIP-based retrieval models} such as BiomedCLIP~\cite{zhang2023biomedclip} and PubMedCLIP~\cite{eslami2023pubmedclip} match images to pre-defined findings via contrastive learning, yielding deterministic outputs but limited expressiveness. \textbf{Generative models} such as LLaVA-Med~\cite{li2023llavamed}, Med-Flamingo~\cite{moor2023medflamingo}, CheXagent~\cite{chen2024chexagent}, RadFM~\cite{wu2023radfm}, InternVL2~\cite{chen2024internvl2}, and Qwen2-VL~\cite{wang2024qwen2vl} produce free-text reports with richer clinical detail but introduce the risk of open-ended hallucination. \textbf{Commercial APIs} such as Gemini~\cite{google2024gemini} and GPT-4V~\cite{yang2024gpt4v} produce fluent output but offer limited transparency about training data and architecture.
\subsection{Hallucination in VLMs}

In the general domain, CHAIR (Caption Hallucination Assessment with Image Relevance)~\cite{rohrbach2018chair} checks whether captione
d objects appear in ground truth. POPE (Polling-based Object Probing Evaluation)~\cite{li2023pope} probes VLMs with yes/no object-presence questions. FaithScore~\cite{jing2024faithscore} decomposes text into atomic claims for verification. In the medical domain, Med-Halt~\cite{umapathi2023medhalt} targets text-only QA without visual grounding, and MedVH~\cite{gu2024medvh} evaluates single models without cross-model comparison or severity. Med-HallMark~\cite{chen2024medhallmark} introduces a hierarchical hallucination taxonomy for medical VLMs but evaluates primarily on VQA tasks without per-finding fabrication/omission analysis or ensemble mitigation. FactCheXcker~\cite{heiman2025factchexcker} addresses measurement-specific hallucinations in CXR report generation through a modular post-hoc correction framework, complementing our focus on diagnostic finding-level hallucination. Recent surveys~\cite{huang2023hallucination} identify fabrication and omission as primary failure modes but do not provide fine-grained medical taxonomies, severity ratings, or concrete mitigation strategies. Table~\ref{tab:comparison} positions our work against these concurrent benchmarks.

\begin{table}[t]
\centering
\caption{Comparison with concurrent medical VLM hallucination benchmarks. Med-HM: Med-HallMark; FactCX: FactCheXcker. Meas.=measurements; Find.=findings; sev.=severity; Hier.=hierarchical.}
\label{tab:comparison}
\footnotesize
\setlength{\tabcolsep}{1.5pt}
\begin{tabular}{lp{1.5cm}p{1.4cm}p{1.5cm}}
\toprule
& \textbf{Med-HM} & \textbf{FactCX} & \textbf{Ours} \\
\midrule
Eval.\ level    & VQA       & Meas.    & Find.+sev. \\
Cross-model     & 3 models  & Single   & 6$\times$3 \\
Per-label       & \ding{55} & Partial  & \ding{51} \\
Severity        & Hier.     & \ding{55}& 3-level \\
Mitigation      & \ding{55} & Post-hoc & Ensemble \\
Risk scoring    & \ding{55} & \ding{55}& ROC/AUC \\
Clin.\ strata   & \ding{55} & \ding{55}& 4 strata \\
Normal img.     & \ding{55} & \ding{55}& \ding{51} \\
\bottomrule
\end{tabular}
\end{table}

\subsection{LLM-as-Judge for Evaluation}

Large language models are now routinely used as automated evaluators, offering a scalable alternative to human annotation. MT-Bench and Chatbot Arena~\cite{zheng2024llmasjudge,chiang2024chatbot} showed that LLM judges approximate human preference rankings with reasonable fidelity. In medical AI, where expert annotators are scarce, this approach is particularly attractive. We adopt domain-specific radiological prompting for structured hallucination detection with type classification and severity rating. To detect same-model bias, we cross-validate using GPT-5.1 as a second independent judge, and we validate both the auto-detection pipeline and the LLM judge against 250 stratified human annotations (Section~\ref{sec:validation}).

\section{Method}

\subsection{Dataset Construction}

We sample 856 frontal chest radiographs from MIMIC-CXR~\cite{johnson2019mimiccxr}, representing 220 unique patients. Images are stratified into four complexity levels using CheXpert labels~\cite{irvin2019chexpert} (14 radiological observations extracted from reports by NLP): \textbf{S1}~(Normal, $n{=}153$): no positive CheXpert findings; \textbf{S2}~(Single, $n{=}212$): exactly one finding; \textbf{S3}~(Multi, $n{=}241$): 2--3 findings; \textbf{S4}~(Complex, $n{=}250$): $\geq$4 findings. The 12 pathological CheXpert labels serve as ground truth, supplemented by original radiologist reports. CheXpert's uncertain labels ($-1$) are stored separately: mentioning an uncertain finding is not penalized as fabrication, and omission counts only confirmed-positive labels. A sensitivity analysis treating uncertain as negative increases rates by only 0.5--1.6~pp. Table~\ref{tab:label_dist} shows the distribution of CheXpert labels across the dataset, with Pleural Effusion (275 images), Lung Opacity (261), and Cardiomegaly (232) being the most prevalent.

\begin{table}[t]
\centering
\caption{CheXpert label distribution across 856 images.}
\label{tab:label_dist}
\small
\begin{tabular}{lcc}
\toprule
\textbf{Label} & \textbf{Positive} & \textbf{Uncertain} \\
\midrule
Pleural Effusion          & 275 & 24 \\
Lung Opacity              & 261 & 15 \\
Cardiomegaly              & 232 & 28 \\
Atelectasis               & 187 & 33 \\
Edema                     & 181 & 63 \\
Pneumonia                 &  78 & 68 \\
Consolidation             &  47 & 22 \\
Lung Lesion               &  33 &  2 \\
Enlarged Cardiomediastinum&  30 & 40 \\
Pneumothorax              &  21 &  9 \\
Pleural Other             &  19 &  2 \\
Fracture                  &  18 &  2 \\
\bottomrule
\end{tabular}
\end{table}

\subsection{Models}

We evaluate six VLMs from distinct architectural families:
\begin{enumerate}
    \item \textbf{BiomedCLIP}~\cite{zhang2023biomedclip}: CLIP-based similarity retrieval (non-generative). Reports CheXpert labels with cosine similarity ${\geq}0.5$ (above-chance threshold); outputs are prompt-invariant.
    \item \textbf{Gemini 3 Flash}~\cite{google2024gemini}: Commercial multimodal API (proprietary). Processes images via a cloud endpoint with undisclosed architecture.
    \item \textbf{GPT-5.1}\footnote{GPT-5.1 refers to the OpenAI model accessed via the \texttt{gpt-5.1} API endpoint. Results reflect this specific checkpoint.}: Commercial multimodal API (OpenAI). Large-scale reasoning model with vision capabilities, accessed via API with temperature~0.
    \item \textbf{InternVL2}~\cite{chen2024internvl2}: Open-source encoder-decoder (6B). Encodes image patches and text separately before fusing.
    \item \textbf{LLaVA-NeXT-Mistral}~\cite{liu2024llavanext}: Open-source LLM-backbone (7B). Projects image patches directly into the language model as token embeddings.
    \item \textbf{Qwen2-VL}~\cite{wang2024qwen2vl}: Open-source (7B). Processes images at native pixel resolution rather than resizing.
\end{enumerate}
Local models use 4-bit quantization on a single A100; all inference uses temperature~0 and max 512 tokens. GPT-5.1 and Gemini 3 Flash use cloud APIs with identical prompts.

\subsection{Query Design}

Each image is evaluated with three clinically motivated query types, each testing a different dimension of VLM capability:

\noindent\textbf{Open-ended report generation:}
\texttt{``Describe all findings visible in this chest X-ray image. Be specific about locations, severity, and notable features.''}
This tests spontaneous hallucination: what a model fabricates or omits without specific prompting.

\noindent\textbf{Targeted visual question answering:}
\texttt{``Is there evidence of [specific finding] in this image? Answer YES or NO first, then provide a brief explanation.''}
The queried finding is sampled from ground-truth positive labels (expected YES, 79\%) or absent labels (expected NO, 21\%), testing binary discrimination and yes/no biases.

\noindent\textbf{Clinical reasoning:}
\texttt{``Provide a differential diagnosis based on this image.''}
This tests multi-step reasoning fidelity, requiring models to synthesize findings into clinical hypotheses. Note that this query may elicit speculation, particularly on normal images where no findings exist.

All prompts append ``State your confidence (0--100\%)''; values are extracted via pattern matching. This yields $856 \times 6 \times 3 = 15{,}408$ total evaluations.

\subsection{Hallucination Detection Pipeline}

\noindent\textbf{Taxonomy.} We define eight hallucination categories across three groups (Table~\ref{tab:taxonomy}), each assigned severity 1~(minor, e.g., over-stating ``mild'' when unspecified), 2~(moderate, e.g., wrong laterality for a real finding), or 3~(clinically dangerous, e.g., fabricating pneumothorax on a normal image).

\begin{table}[t]
\centering
\caption{Eight-category hallucination taxonomy with severity ranges.}
\label{tab:taxonomy}
\small
\setlength{\tabcolsep}{3pt}
\begin{tabular}{llp{3.0cm}c}
\toprule
\textbf{Code} & \textbf{Type} & \textbf{Definition} & \textbf{Sev.} \\
\midrule
A1 & Fabrication         & Finding not present in GT          & 1--3 \\
A2 & Omission            & Missing a GT finding               & 1--3 \\
A3 & Spatial error       & Wrong laterality/location          & 2--3 \\
A4 & Severity distortion & Over/under-stating severity        & 1--2 \\
B1 & Reasoning error     & Incorrect clinical inference       & 2--3 \\
C1 & Template overfitting& Generic template output            & 1--2 \\
C2 & Self-contradiction  & Conflicting statements             & 1--2 \\
C3 & Detail fabrication  & Invented specific details          & 1--3 \\
\bottomrule
\end{tabular}
\end{table}

\noindent\textbf{Layer~1: Auto-detection.} Negation-aware keyword matching against CheXpert labels covers all 15,408 records: the algorithm extracts findings, applies negation detection (e.g., ``no evidence of,'' ``ruled out''), and compares against ground truth to identify fabrications and omissions per finding.

\noindent\textbf{Layer~2: LLM-as-judge.} Applied to 12,840 records (five models) using Gemini as primary judge with 99.8\% success rate. The judge receives CheXpert labels (positive and uncertain), the radiologist report, and the model output, classifying each hallucination by type (A1--C3) and severity (1$=$no clinical impact, 2$=$unnecessary follow-up, 3$=$wrong treatment or missed finding). We cross-validate using GPT-5.1 as a second judge on 2,808 stratified records.

\noindent\textbf{Validation.}
\label{sec:validation}
We validate against 250 stratified human annotations by two co-authors with clinical informatics training (disagreements resolved by discussion). Auto-detection achieves \textbf{F1$=$0.959} (precision$=$0.976, recall$=$0.943) and Cohen's $\kappa{=}$0.870. The LLM judge achieves F1$=$0.907 on 210 paired annotations (TP$=$171, FP$=$34, FN$=$1, TN$=$4). The judge's low $\kappa{=}$0.150 is a base-rate artifact (97.6\% vs.\ 81.9\% positive rates; only 4 true negatives); we report F1 as the primary metric. Inter-judge reliability (2,158 records with both Gemini and GPT-5.1 judges): $\kappa{=}$0.515, 97.4\% agreement, 76.1\% severity-3 concordance. Beyond binary detection, we validate category-level judgments against 175 human-annotated samples: severity ratings agree within ${\pm}1$ on 97.1\%, and hallucination type categories overlap on 91.2\% between the judge and human annotators. GPT-5.1 serves as cross-validation judge for the other five models; to avoid circular evaluation, its own outputs use auto-detection only (A1/A2, no A3--C3), and its severity rates (Table~\ref{tab:rates}, $\dagger$) may be underestimated due to lower auto-detection recall (0.675--0.853).

\section{Results}

\subsection{Overall Hallucination Rates}

Table~\ref{tab:rates} presents two complementary hallucination metrics. \textbf{Auto-detection} (negation-aware keyword matching, full coverage of 15,408 records) reveals rates of 61.9\% [95\% CI: 60.0--63.8] to 82.3\% [80.8--83.8] across models. BiomedCLIP achieves the lowest auto-detected rate (61.9\%), partly because its non-generative architecture limits fabrication on normal images (S1:~7.6\%). GPT-5.1 (71.1\%) and Gemini 3 Flash (71.2\%) are statistically indistinguishable (bootstrap 95\% CI of difference: $[-2.6, +2.3]$~pp) despite GPT-5.1 producing longer responses (1,725 vs.\ 1,084~chars). Qwen2-VL has the highest rate (82.3\%, $\Delta{=}$5.8~pp vs.\ InternVL2 [CI: 3.5--8.0], $p{<}$0.05), exceeding 90\% on single and multi-finding images (S2:~95.0\%, S3:~92.8\%).

\textbf{Severe hallucination rates} (LLM-judge annotated severity-3; severity ratings agree with human annotators within ${\pm}1$ on 97.1\% of paired samples) reveal the clinically critical dimension: 36.4--80.2\% of outputs contain at least one hallucination with direct clinical impact. BiomedCLIP again performs best (36.4\%), while Qwen2-VL (80.2\%) and LLaVA-NeXT (74.7\%) produce severe hallucinations on the majority of evaluations.

The two layers agree well: auto-detection precision against the judge is 0.986--0.999, with recall 0.675--0.853 (subtler errors like reasoning mistakes are missed). We use auto-detection as the primary metric (full 15,408-record coverage) and the LLM judge for severity and type analysis.

\begin{table*}[t]
\centering
\caption{Hallucination rates (\%) by model and diagnostic complexity. \emph{Auto}: negation-aware keyword detection. \emph{Severe}: judge-annotated severity-3. S1=Normal, S2=Single, S3=Multi (2--3), S4=Complex ($\geq$4). \textbf{Bold}: lowest (best); \underline{underline}: highest (worst) per column. $^\dagger$GPT-5.1 severity via auto-detection (recall 0.675--0.853), which may underestimate its true severe rate; GPT-5.1 serves as cross-validation judge for other models.}
\label{tab:rates}
\small
\begin{tabular}{l ccccc ccccc}
\toprule
& \multicolumn{5}{c}{\emph{Auto Detection Rate (\%)}} & \multicolumn{5}{c}{\emph{Severe Hallucination Rate (\%)}} \\
\textbf{Model} & S1 & S2 & S3 & S4 & All & S1 & S2 & S3 & S4 & All \\
\cmidrule(lr){2-6} \cmidrule(lr){7-11}
BiomedCLIP      & \textbf{7.6} & 81.1 & 75.5 & \textbf{65.7} & \textbf{61.9} & \textbf{32.8} & \textbf{42.1} & 37.2 & \textbf{33.0} & \textbf{36.4} \\
Gemini 3 Flash  & \underline{89.1} & 66.5 & 66.7 & 68.7 & 71.2 & 66.7 & 50.6 & \textbf{31.0} & 33.5 & 43.0 \\
GPT-5.1         & 54.0 & 78.8 & 76.2 & 70.0 & 71.1 & 42.9$^\dagger$ & 55.2$^\dagger$ & 49.7$^\dagger$ & 39.5$^\dagger$ & 46.8$^\dagger$ \\
InternVL2       & 70.8 & 85.1 & 76.8 & 72.7 & 76.6 & 73.2 & 81.3 & 65.9 & 67.0 & 71.4 \\
LLaVA-NeXT      & 69.3 & \textbf{66.5} & \textbf{66.7} & 66.8 & 67.1 & \underline{75.8} & 71.9 & 73.6 & 77.6 & 74.7 \\
Qwen2-VL        & 43.4 & \underline{95.0} & \underline{92.8} & \underline{85.3} & \underline{82.3} & 55.2 & \underline{86.6} & \underline{87.4} & \underline{83.0} & \underline{80.2} \\
\bottomrule
\end{tabular}
\end{table*}

\subsection{Hallucination Taxonomy Analysis}

Across all models, we count 44,888 individual hallucination instances spanning eight types (Table~\ref{tab:type_model}). For GPT-5.1, only A1 and A2 are available via auto-detection; A3--C3 type distributions therefore cover five models. Fabrication (\textbf{A1}: 22,560, 50.3\%) and omission (\textbf{A2}: 13,542, 30.2\%) together account for 80.4\%. GPT-5.1 contributes 4,039 fabrications and 1,303 omissions (3.1:1 ratio). Model-specific patterns among the five judge-evaluated models emerge: InternVL2 dominates spatial errors (578/1,299$=$44.5\%), LLaVA-NeXT produces the most template overfitting (818/1,218$=$67.2\%), and BiomedCLIP produces the most omissions relative to its total (2,836/4,927$=$57.6\%) but the fewest fabrications.

\begin{table}[t]
\centering
\caption{Hallucination instances by type and model (BMC=BiomedCLIP, Gem=Gemini 3 Flash, IVL=InternVL2, LLa=LLaVA-NeXT, Qw2=Qwen2-VL). GPT-5.1 A1/A2 ($^\dagger$) from auto-detection; A3--C3 unavailable as GPT-5.1 serves as cross-validation judge.}
\label{tab:type_model}
\small
\setlength{\tabcolsep}{1.5pt}
\begin{tabular}{lrrrrrr|r}
\toprule
\textbf{Type} & \textbf{BMC} & \textbf{Gem} & \textbf{GPT} & \textbf{IVL} & \textbf{LLa} & \textbf{Qw2} & \textbf{Total} \\
\midrule
A1 Fabrication    & 1794 & 6143 & 4039$^\dagger$ & 4309 & 2845 & 3430 & 22560 \\
A2 Omission       & 2836 & 1061 & 1303$^\dagger$ & 2618 & 2864 & 2860 & 13542 \\
A3 Spatial        &   33 &   47 &   --- & 578 &  200 &  441 &  1299 \\
A4 Severity       &  180 &  108 &   --- & 492 &  514 &  594 &  1888 \\
B1 Reasoning      &   32 &  266 &   --- &1041 &  777 &  805 &  2921 \\
C1 Template       &   11 &   28 &   --- & 156 &  818 &  205 &  1218 \\
C2 Contradiction  &    0 &    4 &   --- &  61 &   37 &   47 &   149 \\
C3 Detail fab.    &   41 &  400 &   --- & 264 &  487 &  119 &  1311 \\
\midrule
\textbf{Total}    & 4927 & 8057 & 5342$^\dagger$ & 9519 & 8542 & 8501 & 44888 \\
\bottomrule
\end{tabular}
\end{table}

\begin{figure*}[t]
\centering
\includegraphics[width=\textwidth]{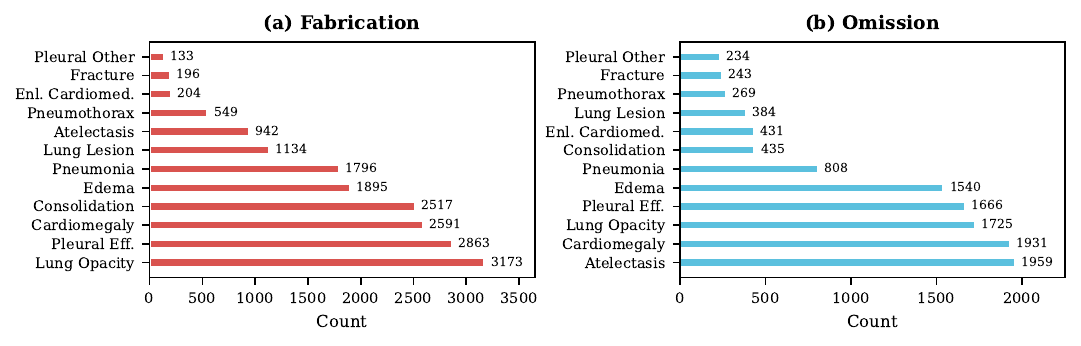}
\caption{Per-label fabrication and omission counts across all models. (a)~Common findings (lung opacity, pleural effusion) are most frequently fabricated; (b)~rare findings (enlarged cardiomediastinum, fracture) are most often omitted.}
\label{fig:perlabel}
\end{figure*}

\noindent\textbf{Per-label fabrication and omission.}
Figure~\ref{fig:perlabel} shows clear asymmetries in how individual CheXpert labels are mishandled. Lung Opacity (29.6\%), Pleural Effusion (28.2\%), and Cardiomegaly (24.1\%) are the most frequently fabricated, all common findings that models appear biased toward generating regardless of image content. Omission rates reveal the inverse pattern: the most commonly missed labels are rare findings such as Enlarged Cardiomediastinum (79.8\%), Fracture (75.0\%), and Pneumothorax (71.2\%). Taken together, \textbf{common findings are over-generated while rare findings are under-detected}, consistent with frequency bias learned from training distributions.

Severity analysis reveals that \textbf{normal radiographs (S1) attract the highest proportion of severe hallucinations} (51.9\% severity-3 vs.\ 27.4--43.8\% for abnormal strata). This may be partly amplified by the clinical reasoning prompt encouraging speculation; however, fabricating findings on a normal image carries maximal clinical risk regardless of prompt effects.

\subsection{Query Type Effects}

Table~\ref{tab:query_cross} breaks down hallucination by query type. Clinical reasoning queries produce the highest rates across all generative models (96.5--98.8\%); targeted VQA produces the lowest (20.8--65.0\%). GPT-5.1 follows this pattern: 96.5\% for clinical reasoning but only 29.8\% for VQA, behind Gemini 3 Flash (20.8\%) and LLaVA-NeXT (20.9\%). Despite generating the longest responses overall (mean 2,751 chars for clinical, 2,083 for open-ended), GPT-5.1 maintains a moderate hallucination rate, which may reflect higher information density per token, though this hypothesis warrants further investigation.

Targeted VQA exposes systematic biases. Given the 79/21\% class imbalance, balanced accuracy (mean of sensitivity and specificity) is more informative: GPT-5.1 (74.2\%) and BiomedCLIP (74.1\%) lead, while Gemini 3 Flash (52.7\%) and LLaVA-NeXT (50.0\%) achieve near-chance because they default to ``YES.'' Qwen2-VL defaults to ``NO'' (specificity 88.8\%, sensitivity 20.8\%).

\begin{table}[t]
\centering
\caption{Auto-detection hallucination rate (\%) and average response length by model and query type.}
\label{tab:query_cross}
\small
\setlength{\tabcolsep}{3pt}
\begin{tabular}{l cc cc cc}
\toprule
& \multicolumn{2}{c}{\textbf{Open}} & \multicolumn{2}{c}{\textbf{VQA}} & \multicolumn{2}{c}{\textbf{Clinical}} \\
\textbf{Model} & \% & Len & \% & Len & \% & Len \\
\cmidrule(lr){2-3} \cmidrule(lr){4-5} \cmidrule(lr){6-7}
BiomedCLIP & 79.1 & 128 & 27.6 &  82 & 79.1 & 128 \\
Gemini 3 Flash & 95.7 &1229 & 20.8 & 187 & 97.2 &1834 \\
GPT-5.1    & 86.9 &2084 & 29.8 & 339 & 96.5 &2752 \\
InternVL2  & 97.2 &1609 & 33.9 & 367 & 98.6 &1399 \\
LLaVA-NeXT & 81.7 &1384 & 20.9 & 396 & 98.8 &1897 \\
Qwen2-VL   & 83.2 &1271 & 65.0 & 207 & 98.8 &1499 \\
\bottomrule
\end{tabular}
\end{table}

BiomedCLIP produces identical rates for open-ended and clinical queries (79.1\%) because its retrieval-based matching is prompt-invariant. Clinical reasoning triggers the most fabrications (8,375 A1) and reasoning errors (1,778 B1), while targeted VQA produces the most omissions proportionally (3,511 A2, 28.7\% of total).

\noindent\textbf{Structured prompting as baseline mitigation.}
Switching from open-ended to targeted VQA lowers hallucination by 18.2--74.9~pp across generative models (mean 55~pp), by constraining outputs to a mean of 299 characters versus 1,876 for clinical reasoning. Where workflows allow binary queries, \textbf{structured VQA is the most cost-effective first-line defense}.

\subsection{Confidence Calibration}
\label{sec:calibration}

\begin{figure*}[t]
\centering
\includegraphics[width=\textwidth]{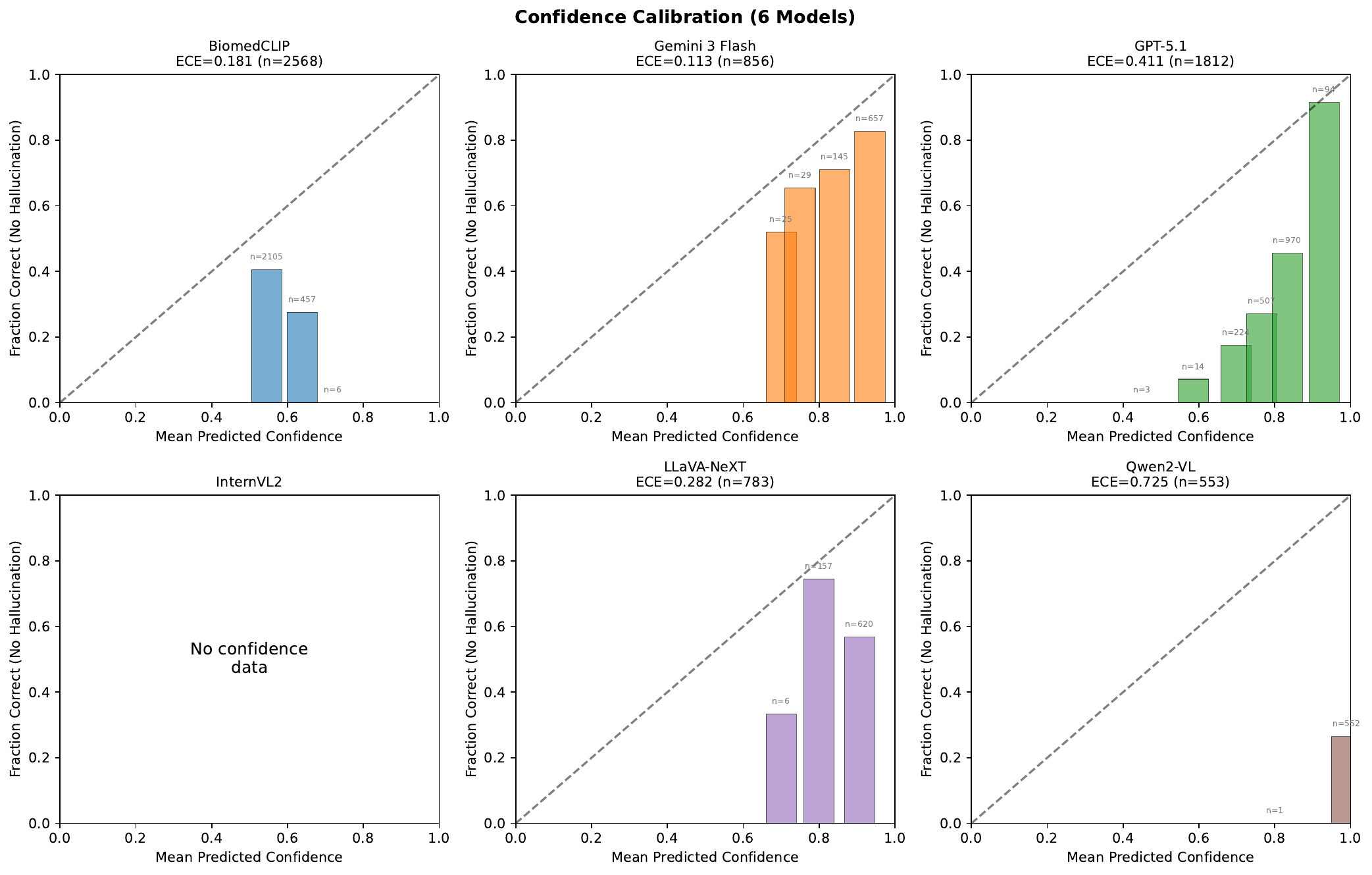}
\caption{Confidence calibration vs.\ hallucination rate. Numbers above bars indicate sample counts per bin. Higher bars at high confidence indicate ``confidently wrong'' predictions. Qwen2-VL shows maximal miscalibration (ECE=0.725).}
\label{fig:calibration}
\end{figure*}

Figure~\ref{fig:calibration} plots confidence calibration for five models (InternVL2 excluded due to insufficient parseable data). Qwen2-VL is the most severely miscalibrated (ECE$=$0.725; ECE measures the gap between stated confidence and actual accuracy): 89\% of its responses report 100\% confidence, yet 73.6\% of these contain hallucinations. GPT-5.1 (ECE$=$0.411) and LLaVA-NeXT (ECE$=$0.282) are also poorly calibrated. BiomedCLIP (ECE$=$0.181) and Gemini 3 Flash (ECE$=$0.113) fare better, though neither is well calibrated in absolute terms. These results indicate that \textbf{model self-reported confidence cannot serve as a reliable safety indicator}.

\subsection{Cross-Model Correlation}

\begin{figure}[t]
\centering
\includegraphics[width=\columnwidth]{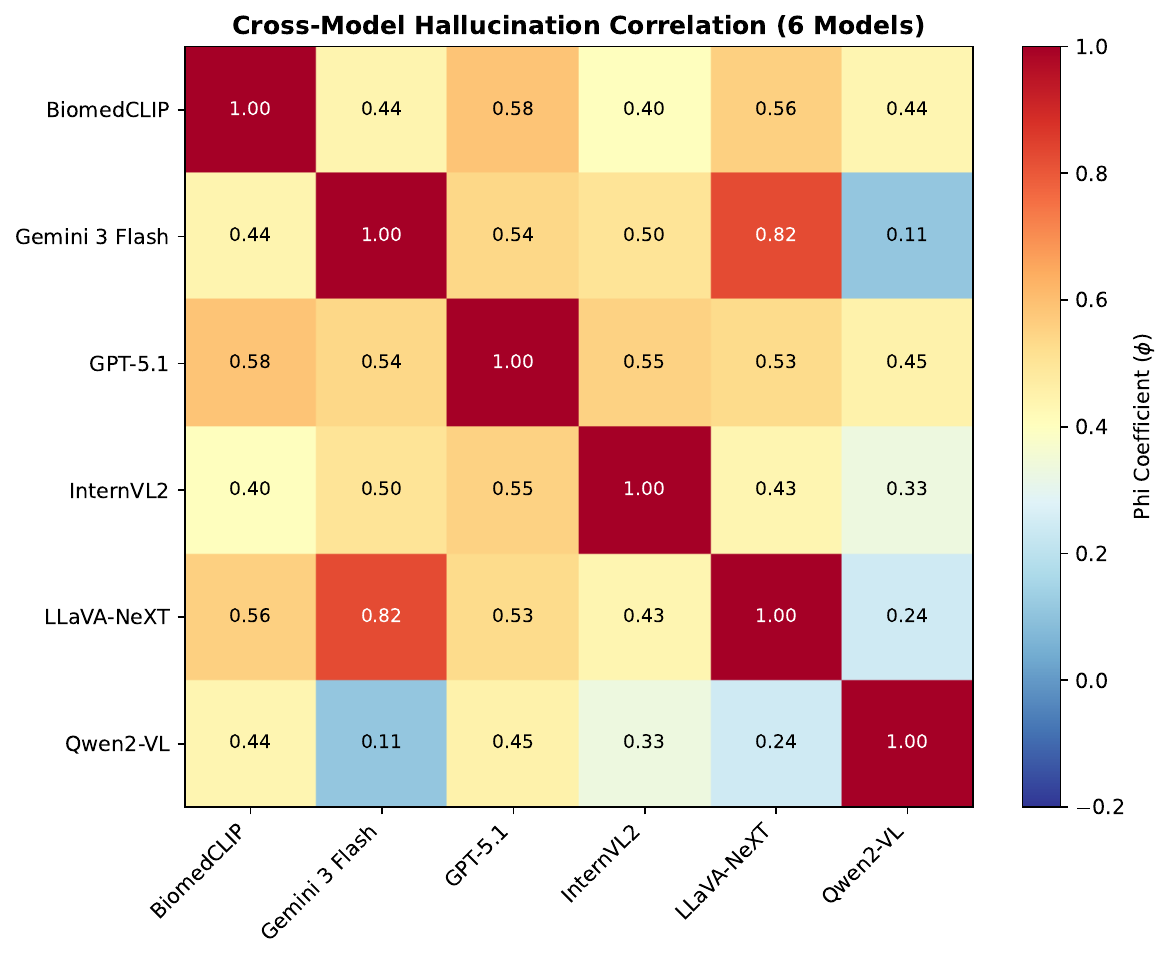}
\caption{Cross-model hallucination correlation (phi coefficients, $\phi$). Correlations range from 0.11 to 0.82; Qwen2-VL shows the most distinctive pattern ($\phi{\leq}0.45$ with all other models).}
\label{fig:correlation}
\end{figure}

We compute phi coefficients ($\phi$; 0$=$independence, 1$=$perfect agreement) between all 15 model pairs (Figure~\ref{fig:correlation}). $\phi$ ranges from 0.114 (Gemini 3 Flash--Qwen2-VL) to 0.824 (Gemini 3 Flash--LLaVA-NeXT), mean 0.46. Qwen2-VL shows the weakest correlations ($\phi{=}$0.11--0.45), indicating a distinctive failure pattern. \textbf{Specific model pairs remain substantially complementary}, sufficient for ensemble-based mitigation.

\subsection{Response Length as Risk Signal}
\label{sec:length_risk}

\begin{figure}[t]
\centering
\includegraphics[width=\columnwidth]{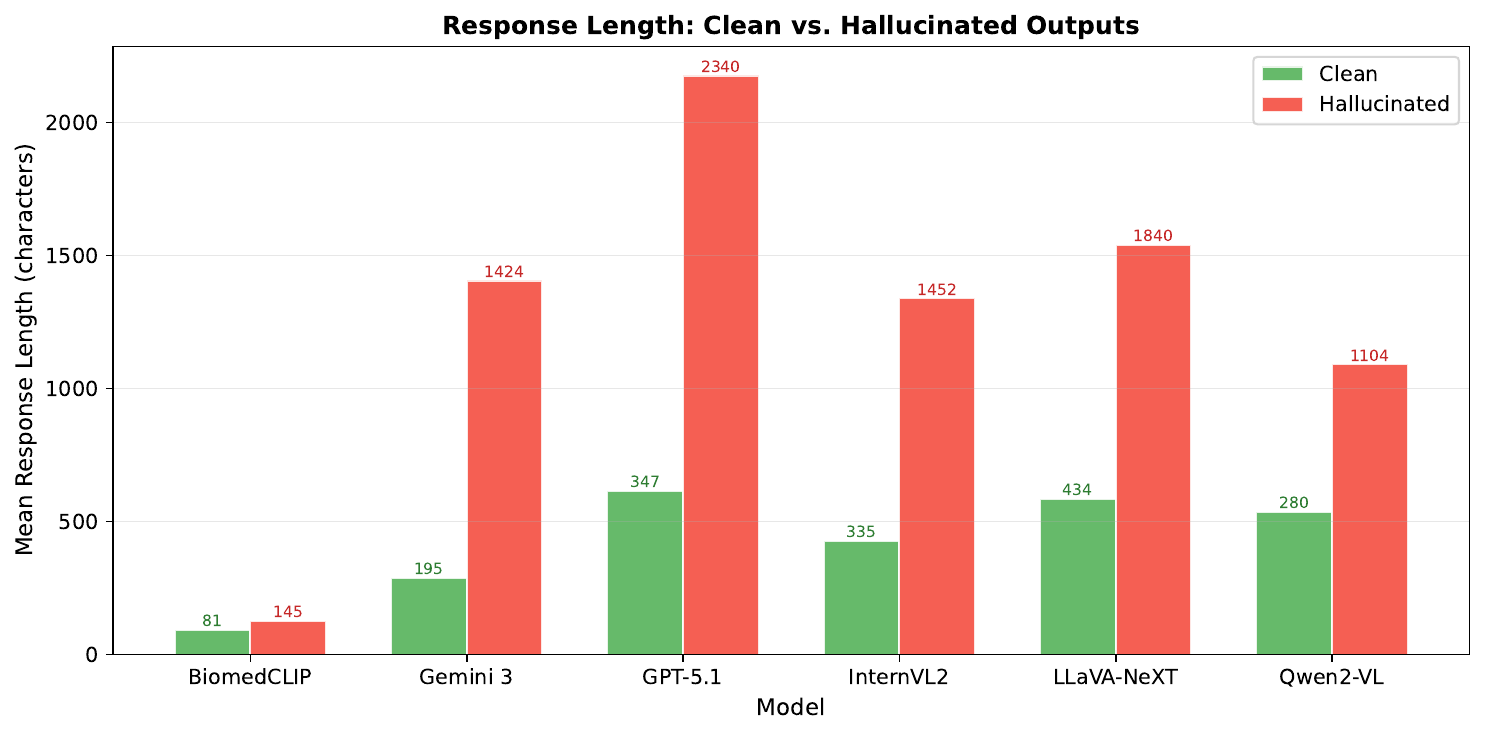}
\caption{Response length distribution for clean vs.\ hallucinated outputs (annotations show median values). Hallucinated responses are consistently longer across all models.}
\label{fig:response_length}
\end{figure}

Response length correlates positively with hallucination (Spearman $\rho{=}0.440$, $p{<}0.001$; point-biserial $r{=}0.472$, the correlation between a binary and a continuous variable). Hallucinated outputs are considerably longer than clean ones across every model (Figure~\ref{fig:response_length}): mean 1,297 vs.\ 402 characters (medians shown in Figure~\ref{fig:response_length}), a $3.2\times$ difference. The gap is widest for Gemini 3 Flash ($4.9\times$: 1,406 vs.\ 287~chars) and InternVL2 ($3.1\times$: 1,338 vs.\ 427~chars).

\begin{figure*}[t]
\centering
\includegraphics[width=\textwidth]{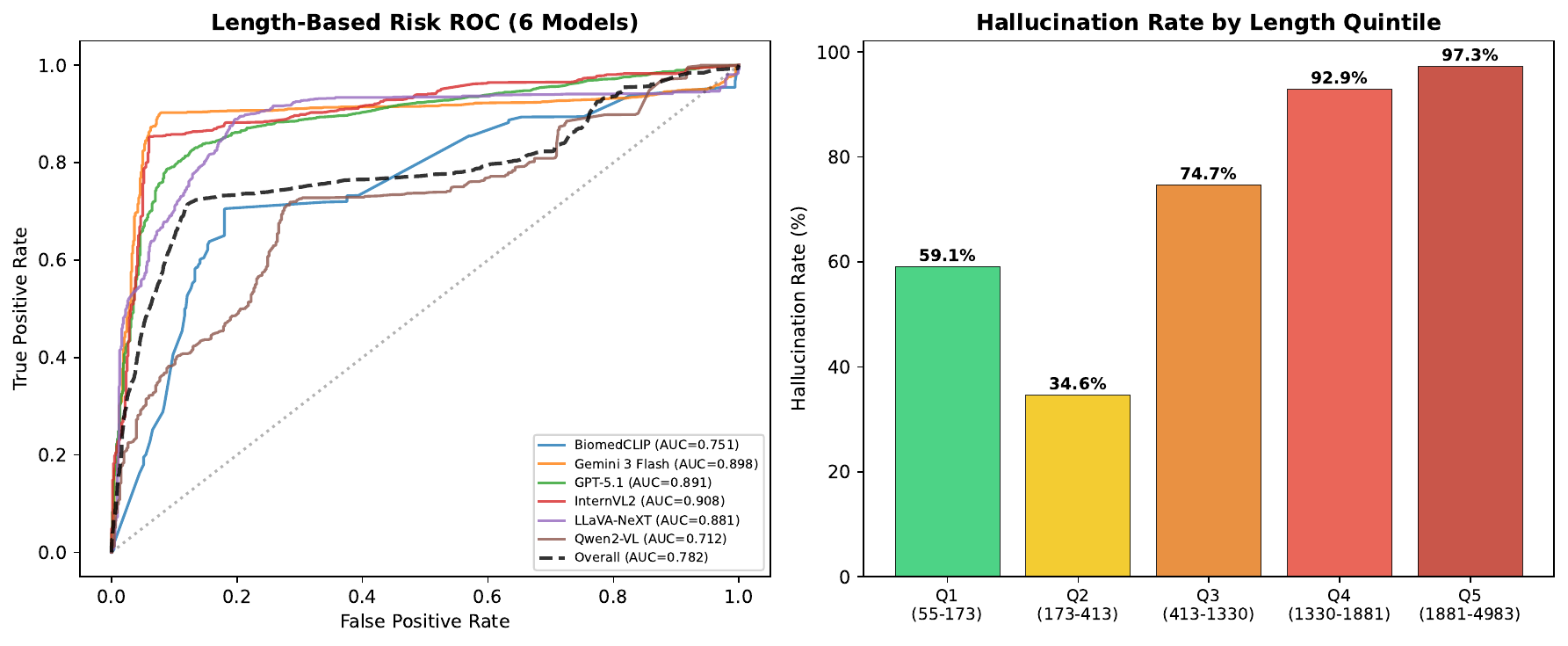}
\caption{Length-based hallucination risk scoring. (a)~ROC curves show per-model AUCs from 0.712 to 0.908. (b)~Hallucination rates by length quintile; rates generally increase with length (Q1 elevated by short BiomedCLIP outputs).}
\label{fig:length_risk}
\end{figure*}

\noindent\textbf{Length-based risk scoring.} We evaluate response length as a lightweight, model-agnostic hallucination risk indicator using ROC analysis (Figure~\ref{fig:length_risk}). Across all models, length achieves AUC$=$0.782 for predicting hallucination presence. Per-model AUCs are substantially higher: InternVL2 (0.908), Gemini 3 Flash (0.898), and LLaVA-NeXT (0.881), suggesting that \textbf{response length is a strong model-specific risk signal}. The optimal threshold at 745 characters (Youden's~$J{=}0.579$, defined as sensitivity + specificity $-$ 1) yields 68.7\% sensitivity and 89.3\% specificity. Hallucination rates generally increase with length: Q1 (55--173~chars) shows 59.1\% (elevated by short BiomedCLIP outputs), dropping to 34.6\% in Q2 (173--413~chars), then rising through Q3 (74.7\%), Q4 (92.9\%), and Q5 (97.3\%, ${\geq}$1,881~chars).

A multi-feature logistic model (length, confidence, query type, stratum) reaches AUC$=$0.858 (5-fold CV), but a length-only logistic model achieves 0.843, so the marginal gain is only $\Delta$AUC${=}$0.015, indicating that \textbf{length alone captures most of the predictable risk}. To assess confounding, we compute within-query-type AUCs (clinical reasoning 0.789, open-ended 0.645, targeted VQA 0.449) and within-stratum AUCs (S1: 0.790, S2: 0.765, S3: 0.798, S4: 0.794). Length is a useful risk signal for clinical reasoning and open-ended queries (AUC$>$0.6) and across all strata (AUC 0.765--0.798), but not for targeted VQA (AUC$=$0.449) where the narrow length range of short responses provides no discriminative power. A simple character-count threshold can thus flag verbose outputs for human review without requiring ground truth or model internals.

\subsection{Ensemble Mitigation}

Despite moderate average correlation, the wide spread in pairwise $\phi$ values (Section~4.5) suggests that ensembles can still exploit complementary errors. We evaluate three strategies at the finding level. Auto-detection provides deterministic binary flags for each (image, query, label) tuple across all 12 CheXpert pathologies and six models.

\noindent\textbf{Simple threshold voting.} As a baseline, we count affirmative mentions across six models and apply voting thresholds $k{\in}\{1,{\ldots},6\}$ (Table~\ref{tab:ensemble}). At $k{\geq}3$, fabrication drops from 0.329 ($k{\geq}1$) to 0.054 (an \textbf{83.6\% reduction}) while omission rises from 0.124 to 0.530, reflecting the inherent fabrication--omission trade-off. The additional sixth model (GPT-5.1) provides a tighter $k{\geq}4$ threshold option.

\noindent\textbf{ECE-weighted voting.} We weight each model's vote by inverse ECE, normalized to sum to one: Gemini 3 Flash receives the highest weight (39.5\%, ECE$=$0.113), Qwen2-VL the lowest (6.2\%, ECE$=$0.725). At threshold 0.5, this achieves the \textbf{best overall F1 (0.563, 95\% CI: 0.55--0.58)}, improving over simple $k{\geq}3$ by 8.7\%, while reducing omission from 0.530 to 0.358 and maintaining fabrication at 0.099.

\noindent\textbf{Label-aware weighted voting.} We assign per-(model, label) weights based on historical precision and specificity via 5-fold CV. This achieves the \textbf{lowest fabrication rate (0.050, an 84.8\% reduction from baseline)}, with the largest improvements for Consolidation ($\Delta$F1$={+}0.135$) and Lung Lesion ($+$0.035). Its omission rate (0.519) remains similar to simple voting, indicating that label-specific weighting primarily targets false positives.

Each strategy suits a different clinical context (Table~\ref{tab:ensemble}): simple voting for transparency, ECE-weighted for screening (recall-oriented), and label-aware for confirmatory settings (precision-oriented). All strategies shift errors along a fabrication--omission trade-off curve; in screening contexts the ECE-weighted strategy (omission$=$0.358) is preferable to $k{\geq}3$ voting (omission$=$0.530). A three-model subset (BiomedCLIP + Gemini 3 Flash + InternVL2) retains 99.9\% of F1 at 50\% cost. Patient-disjoint 5-fold CV confirms robustness ($\Delta$F1$=$${-}$0.015). An oracle selecting the correct model per (image, label) achieves 99.9\% F1, showing large room for learned fusion.

\begin{table}[t]
\centering
\caption{Ensemble mitigation results (6 models). Simple: threshold voting at $k$. ECE-wt.: calibration-weighted. Label-aware: per-label precision-weighted (5-fold CV). Best F1 in \textbf{bold}.}
\label{tab:ensemble}
\small
\begin{tabular}{lccccc}
\toprule
\textbf{Strategy} & \textbf{Fab.} & \textbf{Omis.} & \textbf{Prec.} & \textbf{Rec.} & \textbf{F1} \\
\midrule
Simple $k{\geq}1$   & .329 & .124 & .293 & .876 & .439 \\
Simple $k{\geq}2$   & .149 & .300 & .423 & .700 & .527 \\
Simple $k{\geq}3$   & .054 & .530 & .576 & .470 & .518 \\
Simple $k{\geq}4$   & .008 & .727 & .846 & .273 & .413 \\
Simple $k{\geq}5$   & .000 & .808 &1.000 & .192 & .322 \\
Simple $k{\geq}6$   & .000 & .851 &1.000 & .149 & .259 \\
\midrule
ECE-weighted         & .099 & .358 & .501 & .642 & \textbf{.563} \\
Label-aware CV       & .050 & .519 & .599 & .481 & .534 \\
\bottomrule
\end{tabular}
\end{table}

\section{Discussion}

\noindent\textbf{The normal-image paradox.}
Normal radiographs attract the most severe hallucinations (51.9\% severity-3 vs.\ 27.4--43.8\% for abnormal strata), likely because training corpora over-represent abnormal reports, biasing models toward pathological findings that become pure fabrications on normal images. The clinical reasoning prompt may also encourage speculation on normal studies. Miscalibration compounds this: Qwen2-VL reports 100\% confidence on 89\% of outputs while hallucinating on 80.2\%.

\noindent\textbf{Ensemble potential and limits.}
Cross-model correlations ($\phi{=}$0.11--0.82) allow ensembles to exploit complementary errors ($k{\geq}3$: 83.6\% fabrication reduction; ECE-weighted: best F1$=$0.563). All strategies trade fabrication for omission; the ensemble is a \emph{pre-filter} for human review where false negatives proceed to radiologist interpretation. The gap to oracle (99.9\%) motivates learned fusion; 6$\times$ cost limits broader applicability.

\noindent\textbf{Length as safety trigger.}
Response length provides a model-agnostic risk signal (AUC$=$0.782, up to 0.908) requiring no ground truth. Combined with VQA reducing hallucination by 55~pp, this suggests a layered strategy: VQA for binary queries, length monitoring as a filter, and ensembles for high-stakes interpretations.

\section{Limitations}
{\small
Auto-detection relies on 12 CheXpert NLP-extracted labels (0.29\% of fabrications fall outside these); extraction noise for rare labels (e.g., Fracture $n{=}18$) may inflate omission rates. The LLM judge may introduce same-family bias ($\kappa{=}$0.150 reflects near-uniform positive predictions); GPT-5.1 severity uses auto-detection to avoid circular evaluation. Evaluation covers frontal radiographs from a single institution. Fixed prompt templates are used; prompt sensitivity remains for future work. Open-source models use 4-bit quantization vs.\ full-precision commercial APIs, though gaps are not consistently in one direction. Data are from de-identified MIMIC-CXR-JPG~\cite{johnson2019mimiccxr} under PhysioNet credentialed access; code and data will be released upon acceptance.}

\section{Conclusion}

HalluCXR reveals 61.9--82.3\% hallucination rates across six VLMs; length-based scoring (AUC$=$0.782) and ensembles (84.8\% fabrication reduction) offer practical safety layers for deployment.

{\small
\bibliographystyle{ieeetr}
\bibliography{references}
}

\end{document}